\documentclass[11pt]{article}

\usepackage[preprint]{acl}
\usepackage{times}
\usepackage{latexsym}
\usepackage[T1]{fontenc}
\usepackage[utf8]{inputenc}
\usepackage[expansion=false]{microtype}
\usepackage{inconsolata}
\usepackage{graphicx}
\usepackage{amsmath}
\usepackage{amssymb}
\usepackage{booktabs}
\usepackage{array}
\usepackage{multirow}
\usepackage{xcolor}
\usepackage{algorithm}
\usepackage{algorithmic}
\usepackage{subcaption}
\usepackage{placeins}
\usepackage{enumitem}
\usepackage[most]{tcolorbox}
\newcommand{\caselabel}[1]{\textrm{\textbf{#1}}}
\newcommand{\hlbad}[1]{\colorbox{red!18}{\strut\textbf{#1}~$\times$}}
\newcommand{\hlgood}[1]{\colorbox{green!22}{\strut\textbf{#1}~$\checkmark$}}
\tcbset{
  memtext/.style={enhanced, fonttitle=\small, fontupper=\footnotesize\ttfamily\raggedright, fontlower=\footnotesize\ttfamily\raggedright},
  basebox/.style={memtext, colback=gray!3, colframe=gray!50!black},
  arisebox/.style={memtext, colback=blue!3, colframe=blue!40!black},
  casebox/.style={enhanced, colback=blue!3, colframe=blue!40!black, fonttitle=\small, fontupper=\footnotesize\ttfamily\raggedright, left=4pt, right=4pt, top=3pt, bottom=3pt},
}

\title{ARISE: Practical Self-Evolving Memory for Accurate and Cost-Efficient Enterprise Agents}

\author{
  \textbf{Jiangyun Zhang, Zhou Tang, Kristen Surrao, Zhiyu Chen, Albert Guan, }
  \\
  \textbf{Tor Nitayanont, Yupei Zhang, Roopali Singh, Yunkai Xiao, Shreya Pandit,  }
  \\
  \textbf{Julia Huang, Shayan Ali Akbar, Erwin Cornejo, Yuan Li, Yi Zhang}
  \\
  Amazon
  \\
  \small\texttt{\{zjiangyu, tangzhou, kmsurrao, zhiyuche, xianguan, torpong, yupei, roopalis,}
  \\
  \small\texttt{yunkxiao, pandishr, juliahuw, shayaakb, eccornej, liayuan, yizhngn\}@amazon.com}
}

\begin{document}
\hypersetup{pdftitle={ARISE: Practical Self-Evolving Memory for Accurate and Cost-Efficient Enterprise Agents}, pdfauthor={Maxwell Zhang, Zhou Tang, Kristen Surrao, Zhiyu Chen, Albert Guan, Tor Nitayanont, Yupei Zhang, Roopali Singh, Yunkai Xiao, Shreya Pandit, Julia Huang, Shayan Ali Akbar, Erwin Cornejo, Yuan Li, Yi Zhang}}
\maketitle

\begin{figure}[!b]
\centering
\includegraphics[width=\columnwidth]{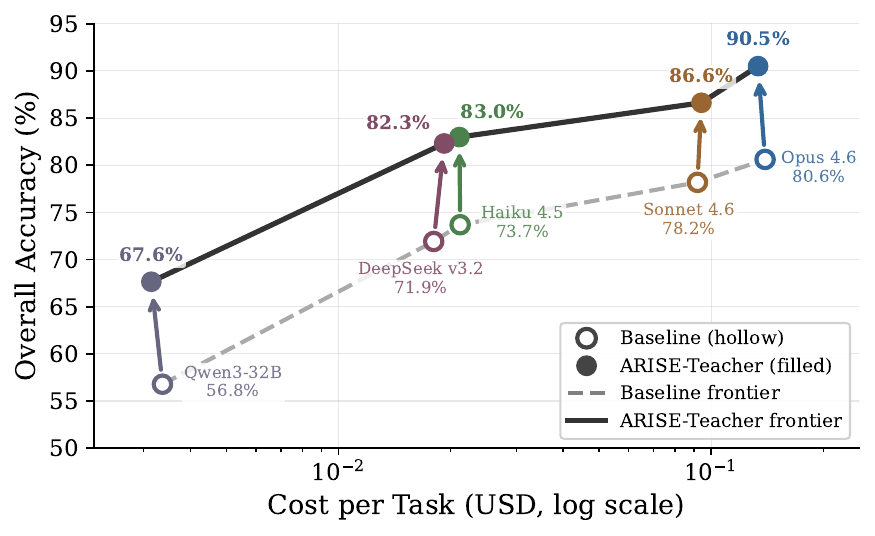}
\caption{Overall accuracy over the 14 domains plotted against SOP-Bench cost per task, per model. The overall accuracy is the 11:3 blend of the suite averages in Table~\ref{tab:main_results}.}
\label{fig:pareto}
\end{figure}

\begin{abstract}
Enterprise LLM agents operate in dynamic environments where foundation models evolve rapidly, workflows update frequently, and deployments need to satisfy accuracy and cost requirements. We propose ARISE (\textbf{A}gent \textbf{R}efinement through \textbf{I}terative \textbf{S}elf-\textbf{E}valuation), a practical memory architecture in which the agent's memory evolves from its own failures without updating model weights, tools, or workflow logic. ARISE uses the model itself or a teacher model to attribute the failures and propose corrections. Then, ARISE keeps the corrections as retrievable micro memory, consolidates them into macro rules, and accepts a rule update only if held-out accuracy rises. Across 14 enterprise-workflow domains from SOP-Bench and $\tau^2$-bench using 5 models, ARISE-Teacher, which uses Claude Opus~4.6 for attribution and consolidation, gains 10.9\,pp over the baseline on SOP-Bench and 5.8\,pp on $\tau^2$-bench. ARISE-Teacher also outperforms GEPA by 4.5\,pp on $\tau^2$-bench and 1.4\,pp on SOP-Bench. With the deployment model as its own attributor, ARISE-Self still gains 10.0\,pp on SOP-Bench. ARISE-Teacher uses 4.0\% fewer model calls and 5.2\% fewer output tokens at comparable per-task cost ($-$1.0\%). 
\end{abstract}

\section{Introduction}
\label{sec:intro}

Enterprise agents run on rapidly evolving LLMs, with new state-of-the-art models appearing monthly. Standard operating procedures (SOPs) and policies change, requiring corresponding updates to agents. Model and workflow changes occur independently, requiring teams to revalidate agents rather than rely on a one-time prompt update. Rebuilding agents for these changes requires development and maintenance effort. Each task has a cost budget, so longer prompts incur cost on every request. Existing methods improve agents from experience through reflection~\citep{shinn2023reflexion}, insight extraction~\citep{zhao2024expel}, or prompt and context evolution~\citep{agrawal2025gepa,zhang2025ace}. Model training~\citep{srivastava2025dte} additionally requires access to model weights. Many enterprise workflows expose verifiable post-task outcomes, such as whether a customer request reaches its intended system state or an SOP execution returns the correct decision.
\begin{figure*}[t]
  \centering
  \includegraphics[width=\textwidth]{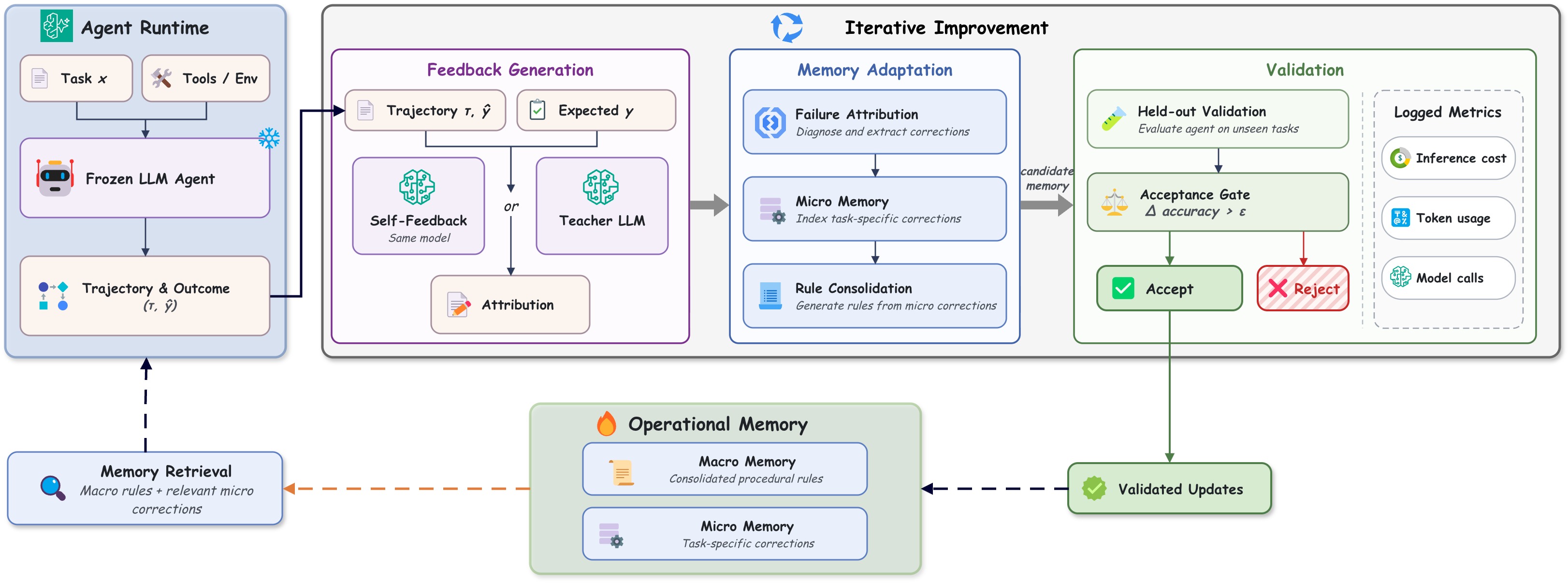}
  \caption{ARISE architecture. The deployed model weights, tools, and workflow logic stay fixed. An offline loop attributes failed tasks using the trace, the output, the post-task outcome and diagnostic feedback, stores micro corrections, consolidates macro rules, and accepts a rule update only if held-out validation accuracy improves.}
  \label{fig:architecture}
\end{figure*}

ARISE is a practical memory architecture that self-evolves while model weights, tools, and workflow logic remain fixed (Figure~\ref{fig:architecture}). The offline loop uses task traces, agent outputs, and post-task feedback to diagnose failures. It collects failed training tasks with their expected outputs or diagnostic feedback, asks an attributor to write scoped corrections, stores them as micro memory, and consolidates them into at most $K$ macro rules. A held-out validation gate accepts a rule update only when accuracy improves. At inference, the agent receives the macro rules and the most similar corrections. A stronger LLM can serve as the teacher, while the deployment model can perform attribution and consolidation itself; the latter setting gains 10.0\,pp on SOP-Bench.

Our contributions are threefold. First, we propose a dual-memory architecture that adapts a deployed agent from its own failures without retraining the model. The bounded rule set keeps the injected memory compact and directly inspectable before deployment. Micro memory preserves corrections, macro memory consolidates general guidance, and a held-out gate controls rule updates. The same loop can be rerun when the workflow or deployment model changes. Second, across 14 enterprise-workflow domains using 5 models, ARISE-Teacher improves over the baseline and GEPA while using 4.0\% fewer model calls and 5.2\% fewer output tokens than the baseline (Tables~\ref{tab:main_results} and~\ref{tab:efficiency_all}). Third, we provide deployment guidance on the sources of the gains, the choice of attributor, and the behavior of the validation gate (\S\ref{sec:guidance}).

\section{Related Work}
\label{sec:related}

\begin{table*}[t]
\centering
\small
\setlength{\tabcolsep}{4pt}
\begin{tabular*}{\textwidth}{@{\extracolsep{\fill}}lcccccc@{}}
\toprule
\textbf{Method} & \textbf{Attribution} & \textbf{Instance-level} & \textbf{Generalized} & \textbf{Summarization} & \textbf{Validation} & \textbf{Injection} \\
\midrule
ExpeL & Self & $\checkmark$ & $\checkmark$ & Insight update & $\times$ & Insights \\
GEPA & Teacher & $\times$ & $\checkmark$ & Evolution & Pareto & Optimized prompt \\
\midrule
ARISE-Self & Self & $\checkmark$ & $\checkmark$ & Bounded rewrite & $\checkmark$ & Top-$k$ + $\leq K$ rules \\
ARISE-Teacher & Teacher & $\checkmark$ & $\checkmark$ & Bounded rewrite & $\checkmark$ & Top-$k$ + $\leq K$ rules \\
\bottomrule
\end{tabular*}
\caption{Design comparison of ARISE with ExpeL and GEPA. \emph{Instance-level}: task-specific experience retrieval. \emph{Generalized}: rules applied across tasks.}
\label{tab:comparison}
\end{table*}

\paragraph{Agent memory and procedural knowledge.}
Agent memory systems preserve information across interactions so later decisions can reuse past context. The Generative Agents architecture~\citep{park2023generative} stores observations, reflections and plans in a memory stream. MemGPT~\citep{packer2023memgpt} manages memory across context and external storage. Mem0~\citep{chhikara2025mem0} and A-MEM~\citep{xu2025amem} organize long-term memories for later retrieval. Voyager~\citep{wang2023voyager} accumulates executable skills from interaction, while MemoryBank~\citep{zhong2024memorybank} preserves long-term user experiences. Beyond persistent storage, procedural memory extracts reusable guidance about how an agent should act from prior trajectories. Reflexion~\citep{shinn2023reflexion}, evaluated in our experiments, turns feedback from failed episodes into verbal reflections reused in later trials. ExpeL~\citep{zhao2024expel} extracts reusable insights from collected experiences and applies them to future tasks. RICP~\citep{sun2024ricp} derives principles from past mistakes, while AutoGuide~\citep{fu2024autoguide} generates context-aware guidelines and AWM~\citep{wang2024awm} induces reusable workflows. Recent systems include Memp~\citep{fang2026memp} for fine-grained and script-like procedures, ReasoningBank~\citep{ouyang2026reasoningbank} for strategies from successful and failed experiences, ReMe~\citep{cao2026reme} for adaptive reuse and refinement, and MNL~\citep{su2026mistake} for performance-gated failure memory. ARISE focuses on an operational combination of retrievable corrections from individual failures, a separately bounded rule set, and a validation gate for rule updates while model weights, tools, and workflow logic remain fixed.

\paragraph{Automated prompt engineering.}
Prompt engineering adapts an LLM by improving the instructions or context supplied at inference. APE~\citep{zhou2022large} generates candidate instructions from demonstrations and selects them using task scores. OPRO~\citep{yang2023large} gives an LLM previous solutions and their scores, then iteratively proposes new instructions. DSPy~\citep{khattab2023dspy} represents multi-stage LM workflows as programs that can be compiled and optimized. MIPRO~\citep{opsahl2024mipro} jointly searches over instructions and demonstrations using Bayesian optimization. ProTeGi~\citep{pryzant2023automatic} derives textual gradients from minibatch errors with bandit selection. EvoPrompt~\citep{guo2024evoprompt} evolves prompts through evolutionary operators and development-set evaluation. GEPA~\citep{agrawal2025gepa} uses reflective mutation and retains a Pareto front of candidates. ACE~\citep{zhang2025ace} evolves context playbooks through reflection. In ARISE, prompt evolution applies to macro rules.

Enterprise workflows often produce verifiable success or failure outcomes that naturally support adaptation and reflection. At deployment scale, additional context from prompts and memory creates recurring serving costs. Extensive changes to the agent harness are also difficult to maintain and govern as models and workflows change. We compare ARISE with the evaluated baselines on accuracy while accounting for serving calls, tokens and inference cost. Table~\ref{tab:comparison} summarizes the design differences.

\section{The ARISE Architecture}
\label{sec:method}

\subsection{Agent Runtime and Feedback}
\label{sec:runtime}

An agent $\mathcal{A}=(p,\mathcal{T},m)$ consists of a system prompt $p$, a tool set $\mathcal{T}$ and a deployment model $m$. Given a task $x$, it produces a ReAct-style trajectory $\tau$~\citep{yao2023react} and an output $\hat{y}$:
\begin{equation}
(\tau,\hat{y}) = \mathcal{A}\bigl(x;\; p \oplus \mathcal{R} \oplus \mathcal{M}(x)\bigr),
\label{eq:runtime}
\end{equation}
where $\mathcal{R}$ is the set of macro rules, $\mathcal{M}(x)$ is the set of micro corrections retrieved for $x$ (\S\ref{sec:retrieval}), and $\oplus$ denotes concatenation. The baseline is $\mathcal{R}=\mathcal{M}(x)=\emptyset$. Each completed training task yields a binary post-task outcome indicating success or failure. For a failed task, feedback $y$ provides diagnostic context for attribution: the expected decision on SOP-Bench or the failed action checks on $\tau^2$-bench. The training data is partitioned into an iteration set $\mathcal{D}_\text{train}$, whose failures drive memory updates, and a validation set $\mathcal{D}_\text{val}$, on which each update is accepted or rejected. The test set $\mathcal{D}_\text{test}$ is used for evaluation only. ARISE modifies only the context in Eq.~\ref{eq:runtime} and leaves $\mathcal{T}$ and $m$ fixed, so it applies to closed-source models. Training runs offline, and serving reads the accepted memory.

\subsection{Failure Attribution}
\label{sec:attributor}

For each failure $j$ in an iteration batch, an attributor LLM writes a scoped correction:
\begin{equation}
c_j = \mathrm{Attr}\bigl(x_j,\, y_j,\, \hat{y}_j,\, \tau_j\bigr).
\label{eq:attr}
\end{equation}
The attributor $\mathrm{Attr}$ can be the deployment model $m$ itself or a stronger teacher model, using the prompt in Figure~\ref{fig:prompts}. Each $c_j$ is a single line \texttt{[WHEN: condition] action}, whose condition restricts the tasks it applies to.

\begin{figure*}[!t]
\centering
\includegraphics[width=0.75\textwidth]{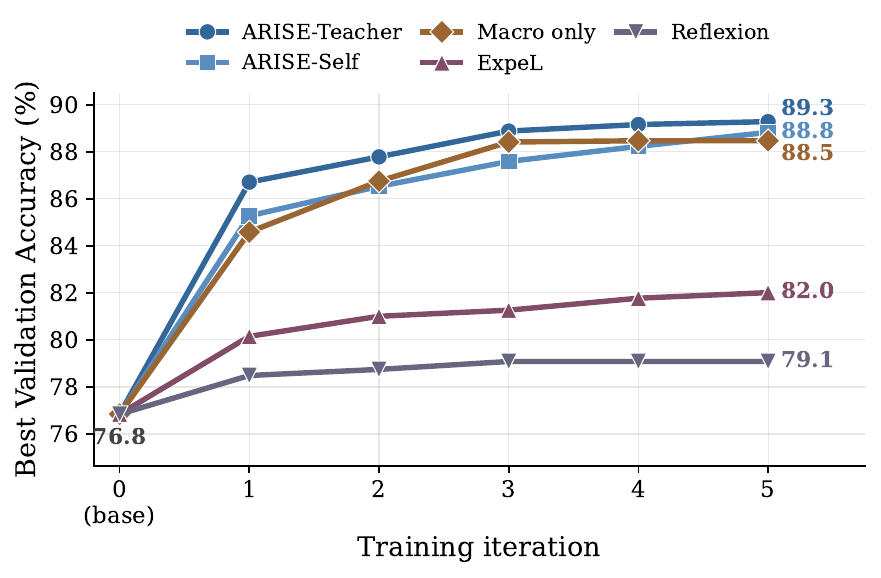}
\caption{Best SOP-Bench validation accuracy so far after each training iteration, averaged over training runs and models. Iteration 0 is the original prompt on the validation set.}
\label{fig:convergence}
\end{figure*}

\begin{table*}[t]
\centering
\scriptsize
\setlength{\tabcolsep}{2.6pt}
\begin{tabular*}{\textwidth}{@{\extracolsep{\fill}}l|cccccc|rrrr|rrrr@{}}
\toprule
\multirow{2}{*}{\textbf{Model}} & \multicolumn{6}{c|}{\textbf{Accuracy (\%)}} & \multicolumn{4}{c|}{\textbf{ARISE-Self over Baseline}} & \multicolumn{4}{c}{\textbf{ARISE-Teacher over Baseline}} \\
\cmidrule(lr){2-7} \cmidrule(lr){8-11} \cmidrule(lr){12-15}
& \textbf{Base} & \textbf{ExpeL} & \textbf{Refl.} & \textbf{GEPA} & \textbf{Self} & \textbf{Teacher} & \textbf{$\Delta$Acc.} & \textbf{$\Delta$Tok.} & \textbf{$\Delta$Calls} & \textbf{$\Delta$Cost} & \textbf{$\Delta$Acc.} & \textbf{$\Delta$Tok.} & \textbf{$\Delta$Calls} & \textbf{$\Delta$Cost} \\
\midrule
\multicolumn{15}{@{}l}{\textit{SOP-Bench (11 domains; one test pass per trained configuration)}} \\
\midrule
Qwen3-32B       & 65.9 & 68.2 & 67.0 & 72.8 & \textbf{78.4} & \underline{78.1} & +12.5 & $-$7.4 & $-$8.9 & $-$7.6 & +12.2 & $-$6.2 & $-$8.6 & $-$6.6 \\
DeepSeek v3.2   & 76.6 & 78.3 & 76.6 & \textbf{89.2} & 84.3 & \underline{88.6} & +7.7 & +0.8 & $-$2.3 & +0.8 & +12.0 & +6.7 & $-$0.9 & +6.4 \\
Haiku 4.5       & 79.2 & 74.7 & 78.8 & \textbf{88.3} & 87.2 & \underline{88.0} & +8.0 & +5.7 & +0.7 & +3.6 & +8.8 & +0.8 & $-$2.6 & $-$0.7 \\
Sonnet 4.6      & 79.2 & 75.0 & 80.2 & 87.3 & \textbf{89.4} & \underline{89.2} & +10.2 & +2.2 & $-$2.2 & +1.1 & +10.0 & +3.6 & $-$2.1 & +2.4 \\
Opus 4.6        & 81.7 & 82.5 & 82.7 & 92.4 & \textbf{93.1} & \textbf{93.1} & +11.4 & $-$4.4 & $-$7.3 & $-$4.3 & +11.4 & $-$4.4 & $-$7.3 & $-$4.3 \\
\cmidrule{1-15}
\textit{Avg}    & 76.5 & 75.7 & 77.1 & 86.0 & \underline{86.5} & \textbf{87.4} & +10.0 & $-$0.6 & $-$4.0 & $-$1.3 & +10.9 & +0.1 & $-$4.3 & $-$0.6 \\
\midrule
\multicolumn{15}{@{}l}{\textit{$\tau^2$-bench (3 domains)}} \\
\midrule
Qwen3-32B       & 23.3 & 24.6 & 21.7 & 22.9 & \underline{27.2} & \textbf{29.3} & +3.9 & $-$9.0 & $-$6.5 & $-$9.1 & +6.0 & $-$11.7 & $-$10.3 & $-$11.9 \\
DeepSeek v3.2   & 54.6 & 55.4 & \underline{56.9} & 56.5 & 55.3 & \textbf{59.2} & +0.6 & $-$5.0 & $-$3.1 & $-$5.2 & +4.6 & $-$2.2 & $-$2.5 & $-$2.3 \\
Haiku 4.5       & 53.5 & 53.8 & 54.2 & 55.0 & \underline{56.7} & \textbf{64.6} & +3.1 & $-$4.5 & $-$2.7 & $-$4.4 & +11.1 & $-$3.1 & $-$3.7 & $-$3.1 \\
Sonnet 4.6      & 74.6 & 67.5 & 73.1 & \underline{76.7} & 75.6 & \textbf{77.3} & +1.0 & +0.1 & $-$0.4 & $-$0.3 & +2.7 & +3.2 & +0.6 & +2.8 \\
Opus 4.6        & 76.5 & 76.5 & 73.6 & 77.5 & \textbf{80.9} & \textbf{80.9} & +4.4 & +1.8 & +0.4 & +1.7 & +4.4 & +1.8 & +0.4 & +1.7 \\
\cmidrule{1-15}
\textit{Avg}    & 56.5 & 55.5 & 55.9 & 57.8 & \underline{59.1} & \textbf{62.3} & +2.6 & $-$3.3 & $-$2.5 & $-$3.4 & +5.8 & $-$2.4 & $-$3.1 & $-$2.6 \\
\midrule
\multicolumn{15}{@{}l}{\textit{Overall (14 domains)}} \\
\midrule
\textbf{Avg}    & 72.2 & 71.4 & 72.5 & 80.0 & \underline{80.6} & \textbf{82.0} & +8.4 & $-$1.2 & $-$3.7 & $-$1.7 & +9.8 & $-$0.4 & $-$4.0 & $-$1.0 \\
\bottomrule
\end{tabular*}
\caption{Accuracy (\%) and changes from baseline (equal-domain averages, overall weights domains 11:3). ARISE values average three runs. On Opus~4.6 the two settings coincide, so the ARISE-Self values repeat the ARISE-Teacher values. GEPA SOP-Bench: per-task re-evaluation. $\tau^2$-bench evaluation trials are described in \S\ref{sec:setup}. $\Delta$Acc.\ is in pp from unrounded values. $\Delta$Tok.\ (input+output), $\Delta$Calls and $\Delta$Cost are in \%. Bold: best. Underline: second.}
\label{tab:main_results}
\end{table*}

\subsection{Memory Adaptation}
\label{sec:summarizer}
\label{sec:memory}

\paragraph{Micro memory.} The corrections generated by $\mathrm{Attr}$ form micro memory and are retrieved by embedding similarity to the incoming task.

\paragraph{Rule consolidation.} After each batch, the same model consolidates the new corrections and current rules $\mathcal{R}$ into two candidates: an append candidate that extends the current rules and a rewrite candidate built from the new corrections. Every rule carries a \texttt{[WHEN:]} scope. Both candidates are capped at $K$ rules. When the append candidate exceeds $K$, existing rules are retained first. Appendix~\ref{app:prompts} gives the two consolidation prompts.

\subsection{Validation Gate}
\label{sec:validator}

With $\mathrm{Acc}_v$ the accuracy on $\mathcal{D}_\text{val}$ and $\mathcal{M}$ the micro store, let $\mathcal{D}_{\mathrm{val}}^\mu\subseteq\mathcal{D}_\text{val}$ be the fixed subset for validating micro retrieval and $\mathrm{Acc}_{v,\mu}(\mathcal{R}{+}\mathcal{M})$ its macro-plus-micro accuracy. Algorithm~\ref{alg:arise} applies one gate per memory level:
\begin{align}
    \mathrm{Acc}_v(\mathcal{R}^{\star}) &> \mathrm{Acc}_v(\mathcal{R}) + \epsilon_m \label{eq:macro_gate} \\
    \mathrm{Acc}_{v,\mu}(\mathcal{R}{+}\mathcal{M}) &\geq \mathrm{Acc}_v(\mathcal{R}) - \epsilon_\mu \label{eq:micro_gate} \\
    \epsilon_m &= \max\bigl(\epsilon_0,\; C / |\mathcal{D}_\text{val}|\bigr) \label{eq:eps}
\end{align}
Each candidate that differs from $\mathcal{R}$ is evaluated on $\mathcal{D}_\text{val}$, and the better one $\mathcal{R}^{\star}$ is accepted only if it clears Eq.~\ref{eq:macro_gate}, whose margin widens for small validation sets (Eq.~\ref{eq:eps}). Training stops after $P$ consecutive iterations without an accepted update. After training, if both stores are nonempty, the micro gate screens retrieval by comparing macro only accuracy on the full validation set with macro plus micro accuracy on a fixed subset. Because the scores use different task sets, the gate does not isolate the effect of micro retrieval. Accuracy is the only acceptance criterion.

\begin{algorithm}[t]
\caption{ARISE loop.}
\label{alg:arise}
\footnotesize
\begin{algorithmic}[1]
\REQUIRE base prompt $p^{(0)}$; iteration set $\mathcal{D}_{\mathrm{train}}$; validation sets $\mathcal{D}_{\mathrm{val}},\mathcal{D}_{\mathrm{val}}^\mu$; $k_{\max},K,B,P,\epsilon_m$ (Eq.~\ref{eq:eps}), $\epsilon_\mu$
\STATE $\mathcal{R},\mathcal{M}\leftarrow\emptyset$; $a\leftarrow\mathrm{Acc}_v(\mathcal{R})$; stale $\leftarrow0$
\FOR{$k=1,\ldots,k_{\max}$}
    \STATE \textbf{if} stale $\geq P$ \textbf{then break}
    \STATE run batch $\mathcal{B}_k\subseteq\mathcal{D}_{\mathrm{train}}$, $|\mathcal{B}_k|\leq B$, with $p^{(0)}\!\oplus\!\mathcal{R}$
    \STATE $F_k\leftarrow$ failed tasks in $\mathcal{B}_k$
    \STATE \textbf{if} $F_k=\emptyset$: stale $\leftarrow$ stale$+1$; \textbf{continue}
    \STATE $\Delta\mathcal{M}^{(k)}\leftarrow\mathrm{Attribute}(F_k)$ using Eq.~\ref{eq:attr}; $\mathcal{M}\leftarrow\mathcal{M}\cup\Delta\mathcal{M}^{(k)}$
    \STATE $\mathcal{R}_{\mathrm{add}}\leftarrow\mathrm{Append}(\mathcal{R},\Delta\mathcal{M}^{(k)},K)$; $\mathcal{R}_{\mathrm{rw}}\leftarrow\mathrm{Rewrite}(\Delta\mathcal{M}^{(k)},K)$
    \STATE $\mathcal{C}_k\leftarrow\{\mathcal{R}_{\mathrm{add}},\mathcal{R}_{\mathrm{rw}}\}\setminus\{\mathcal{R}\}$
    \STATE \textbf{if} $\mathcal{C}_k=\emptyset$: stale $\leftarrow$ stale$+1$; \textbf{continue}
    \STATE $\mathcal{R}^{\star}\leftarrow\arg\max_{\widetilde{\mathcal{R}}\in\mathcal{C}_k}\mathrm{Acc}_v(\widetilde{\mathcal{R}})$
    \IF{$\mathrm{Acc}_v(\mathcal{R}^{\star})>a+\epsilon_m$}
        \STATE $\mathcal{R},a\leftarrow\mathcal{R}^{\star},\mathrm{Acc}_v(\mathcal{R}^{\star})$; stale $\leftarrow0$
    \ELSE
        \STATE stale $\leftarrow$ stale$+1$
    \ENDIF
\ENDFOR
\STATE use\_micro $\leftarrow(\mathcal{R}\neq\emptyset)\wedge(\mathcal{M}\neq\emptyset)\wedge[\mathrm{Acc}_{v,\mu}(\mathcal{R}{+}\mathcal{M})\geq\mathrm{Acc}_v(\mathcal{R})-\epsilon_\mu]$
\RETURN $\mathcal{R},\mathcal{M},$ use\_micro
\end{algorithmic}
\end{algorithm}

\subsection{Retrieval at Inference}
\label{sec:retrieval}

Macro rules are injected into every prompt. If the micro gate passed, the task input is embedded and up to $k$ corrections whose cosine similarity exceeds threshold $\theta$ form $\mathcal{M}(x)$ in Eq.~\ref{eq:runtime}. The injected context is bounded at $K$ rules and $k$ corrections.

\subsection{Implementation}
\label{sec:prompts}
Figure~\ref{fig:prompts} shows the two prompts. The attributor writes one scoped correction per failed training task. The serving prompt adds the macro rules and the retrieved corrections to the system message. The task input and the tools stay unchanged. Appendix~\ref{app:implementation} gives the consolidation prompts and the configuration.
\begin{figure}[!t]
\begin{tcolorbox}[basebox, title={\textbf{Attributor prompt}}, after skip=6pt]
You attempted a task and failed.\\[4pt]
Task input (key fields): \{task\_input\}\\
Expected output: \{expected\}\\
Your output: \{predicted\}\\
Your reasoning trace (partial): \{trace\_summary\}\\[4pt]
Write a SHORT, SPECIFIC rule that would prevent this exact failure. Format:\\
{[}WHEN: <specific condition from this task>{]} <what to do differently>\\[4pt]
Be SPECIFIC: mention exact field values, thresholds, or tool outputs you see in the trace. NOT generic advice. One line only.
\end{tcolorbox}
\begin{tcolorbox}[basebox, title={\textbf{Serving prompt}}]
\textrm{\textit{System message}}\\
<instructions>\\
You are an agent that follows Standard Operating Procedures precisely.\\
Use the available tools to gather information, then make a final decision.\\[4pt]
\#\# Important rules\\
\{accepted macro rules, at most $K$\}\\[4pt]
\#\# Reference notes (from similar past tasks --- use ONLY if directly relevant, ignore otherwise)\\
- \{retrieved correction, at most $k$\}\\
</instructions>\\
<sop>\\
\{SOP document\}\\
</sop>\\[4pt]
\textrm{\textit{User message}}\\
Process this task following the SOP:\\
\{task input fields\}\\
Use the available tools, then provide your final output in XML format:\\
<\{decision field\}>your\_answer</\{decision field\}>
\end{tcolorbox}
\caption{Prompts that write and serve the memory. The reference notes appear only when the micro gate passed. On $\tau$, \{expected\} holds the failed action checks, and the two memory sections are added to the policy prompt.}
\label{fig:prompts}
\end{figure}

\section{Evaluation}
\label{sec:experiments}

\subsection{Setup}
\label{sec:setup}

\paragraph{Benchmarks.} SOP-Bench~\citep{sopbench2025} gives each task an SOP document, inputs and deterministic tools. We use 11 domains (1,813 tasks), split each by row into 70\% training and 30\% testing, and score exact match on the decision field. $\tau^2$-bench~\citep{barres2025tau2}, which extends $\tau$-bench~\citep{yao2024tau}, provides multi-turn airline, retail and telecom conversations. We evaluate its official task files and refer to this three-domain evaluation as $\tau$. For both suites, training data is split 70/30 into iteration and validation sets.

\paragraph{Models and cost.} We use 5 models from 3 providers in Amazon Bedrock: Claude Opus~4.6, Sonnet~4.6 and Haiku~4.5, DeepSeek~v3.2, and Qwen3-32B. Agent, attribution and consolidation calls use temperature 0. Costs use the API's list prices per 1M input/output tokens (US\$): Opus 5.50/27.50, Sonnet 3.30/16.50, Haiku 1.10/5.50, DeepSeek 0.62/1.85 and Qwen 0.15/0.60.

\paragraph{Methods.} \textbf{Baseline} uses the original prompt. \textbf{ExpeL}~\citep{zhao2024expel} and \textbf{Reflexion}~\citep{shinn2023reflexion} are reimplemented with the deployment model. \textbf{GEPA}~\citep{agrawal2025gepa} uses Opus~4.6 as reflection model (temperature 1.0). \textbf{Macro} uses a short diagnosis attributor followed by ARISE consolidation and validation, without storing or retrieving micro memory. In \textbf{ARISE-Teacher}, Opus~4.6 attributes and consolidates. In \textbf{ARISE-Self}, the deployment model attributes its own failures and consolidates the rules, with no stronger LLM. Each ARISE setting is trained three times per domain and model (Appendix~\ref{app:config}).

\paragraph{Hyperparameters.} All models and domains use $K{=}5$, top-$k{=}5$, $\theta{=}0.6$, a maximum batch size of $B{=}30$, $k_{\max}{=}5$, $P{=}3$, $\epsilon_0{=}2$\,pp, $C{=}50$, $\epsilon_\mu{=}5$\,pp (sensitivity in \S\ref{sec:sensitivity}).

\paragraph{Metrics.} We report task success rate, or accuracy, defined as the percentage of test tasks solved by an agent. An SOP-Bench task is solved when the decision field exactly matches the expected label. For $\tau$, we use the action evaluator implemented in the released $\tau^2$-bench package. On $\tau$, Baseline, ExpeL, Reflexion, GEPA, Macro and ARISE-Teacher use four test trials. Their accuracy is mean trial success (pass\textasciicircum1), and pass\textasciicircum4 is the percentage of tasks solved in all four trials. ARISE-Self uses one test pass from each of three runs in the main and full accuracy tables, except Opus~4.6, which reuses the identical Teacher evaluations. The compact reliability table uses four trials from its first run. Suite averages weight domains equally, and the overall average weights the 11 SOP-Bench and 3 $\tau$ domains 11:3. For efficiency we report input and output tokens, model calls and dollar cost per task. Significance tests use a stratified paired bootstrap over test tasks and training runs (Appendix~\ref{app:stats}).

\subsection{Main Results}
\label{sec:results}
In Table~\ref{tab:main_results}, ARISE-Teacher has the highest average on both suites, with 87.4\% on SOP-Bench and 62.3\% on $\tau$. It gains 10.9\,pp over the baseline on SOP-Bench, 5.8\,pp on $\tau$ and 9.8\,pp over all 14 domains. The SOP-Bench gain holds for every model, from 8.8\,pp on Haiku~4.5 to 12.2\,pp on Qwen3-32B. ARISE-Self gains 10.0\,pp on SOP-Bench, while ExpeL and Reflexion stay near the baseline at $-$0.8 and +0.6\,pp. ARISE-Teacher outperforms GEPA by 1.4\,pp on SOP-Bench, 4.5\,pp on $\tau$ and 2.0\,pp overall. Opus gains more on SOP-Bench, while Haiku gains more on $\tau$, suggesting workflow type matters more than model scale. Overall gains span 8.4 to 10.9\,pp across all 5 models and 3 providers (Figure~\ref{fig:pareto}).

\begin{table}[!htb]
\centering
\footnotesize
\setlength{\tabcolsep}{2.5pt}
\begin{tabular}{@{}ll|cccc@{}}
\toprule
\textbf{Domain} & \textbf{Metric} & \textbf{Base} & \textbf{GEPA} & \textbf{Self} & \textbf{Teacher} \\
\midrule
\multirow{2}{*}{Airline}
 & pass\textasciicircum1 & 55.2 & 55.2 & \underline{56.2} & \textbf{58.8} \\
 & pass\textasciicircum4 & \underline{41.0} & 38.0 & 40.3 & \textbf{44.3} \\
\midrule
\multirow{2}{*}{Retail}
 & pass\textasciicircum1 & 48.1 & 46.1 & \textbf{49.1} & \underline{48.5} \\
 & pass\textasciicircum4 & 34.0 & 28.0 & \textbf{39.8} & \underline{38.0} \\
\midrule
\multirow{2}{*}{Telecom}
 & pass\textasciicircum1 & 66.1 & 71.8 & \underline{72.2} & \textbf{79.5} \\
 & pass\textasciicircum4 & 44.5 & 52.0 & \underline{55.8} & \textbf{60.7} \\
\bottomrule
\end{tabular}
\caption{Reliability on $\tau$. ARISE-Self values use four trials from its first training run. ARISE-Teacher values average metrics from four trials computed separately within three training runs.}
\label{tab:tau_consistency}
\end{table}

\begin{figure*}[!t]
\centering
\begin{minipage}[t]{0.48\textwidth}
  \vspace{0pt}\centering
  \includegraphics[width=\textwidth]{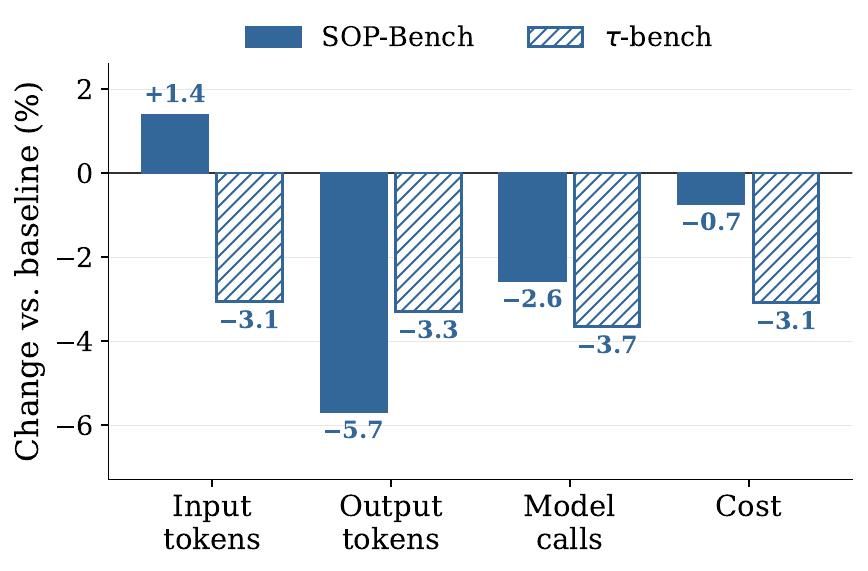}
  \captionof{figure}{Per-task changes in input tokens, output tokens, model calls and inference cost for Haiku~4.5 with ARISE-Teacher relative to the baseline. The legend uses $\tau$-bench for the $\tau$ evaluation.}
  \label{fig:haiku_cost}
\end{minipage}\hfill
\begin{minipage}[t]{0.48\textwidth}
  \vspace{0pt}\centering
  \includegraphics[width=\textwidth]{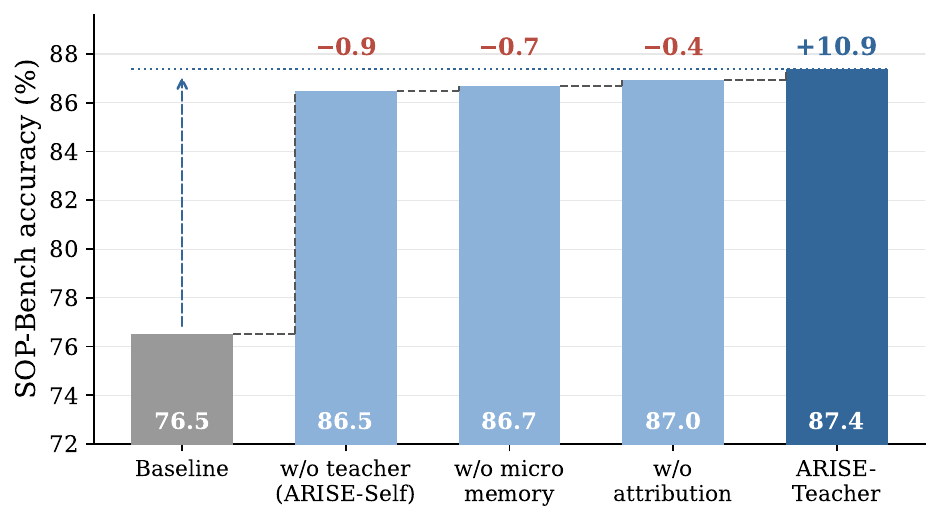}
  \captionof{figure}{Component variants of ARISE-Teacher on SOP-Bench (5-model average accuracy). Labels show each variant's difference from ARISE-Teacher, while +10.9 marks ARISE-Teacher's gain over the baseline. Details: Appendix~\ref{app:ablation}.}
  \label{fig:ablation}
\end{minipage}
\end{figure*}
\subsection{Reliability}
\label{sec:reliability}

On $\tau$ (Table~\ref{tab:tau_consistency}), ARISE-Teacher has the highest pass\textasciicircum4 on airline and telecom, and ARISE-Self on retail. GEPA's telecom pass\textasciicircum1 is 71.8, but its pass\textasciicircum4 is 52.0, against 60.7 for ARISE-Teacher. On retail, ARISE-Teacher raises pass\textasciicircum4 by 4.0\,pp. On telecom, it raises pass\textasciicircum4 for every model, by 10.8 to 26.7\,pp (Appendix~\ref{app:tau_consistency}), and ARISE-Self raises it from 44.5 to 55.8.

\subsection{Component Analysis and Cost Efficiency}
\label{sec:efficiency}
\label{sec:training_cost}
\label{sec:ablation}

\paragraph{Serving cost.} Cost includes model token charges but excludes embedding, storage and retrieval. ARISE adds about 1,340 characters of memory per SOP-Bench task. With ARISE-Teacher, agents nevertheless make 4.0\% fewer model calls over the 14 domains and write 4.9\% fewer output tokens on SOP-Bench and 6.4\% fewer on $\tau$, corresponding to a 5.2\% reduction over the 14 domains. ARISE-Teacher's per-task inference cost changes by $-$1.0\% over the 14 domains (Table~\ref{tab:main_results}). Fewer calls and shorter outputs offset the added memory context, keeping same model inference cost near the baseline while improving average accuracy across both benchmarks and five models. Figure~\ref{fig:haiku_cost} shows the per-task changes for Haiku~4.5.

\paragraph{Component variants.} Figure~\ref{fig:ablation} compares ARISE-Teacher with targeted variants. With the accepted rules fixed, disabling micro retrieval gives 86.7\%, retaining 10.2\,pp of the 10.9\,pp gain. ARISE-Self gives 86.5\%, 0.9\,pp lower, while removing attribution gives 87.0\%, 0.4\,pp lower. Attribution shortens stored entries from 673 to 281 characters and lowers inference cost by 6.0\% when corrections are retrieved.

\paragraph{Feedback variants.} Two further ARISE-Teacher arms change only the feedback, with three training runs each. With pass/fail feedback only, the attributor sees that a task failed but not its expected output, and SOP-Bench accuracy is 79.6\%, 3.1\,pp above the baseline. When 20\% of the training tasks carry incorrect expected labels, accuracy is 83.7\%, 7.2\,pp above the baseline. ARISE-Teacher is 7.8\,pp and 3.7\,pp above these arms. Expected outputs carry most of the training signal, and the loop keeps about two thirds of its gain under 20\% label noise.

\paragraph{Adaptation cost.} Each adaptation run for a model and domain has an estimated model token cost. An ARISE-Teacher run averages \$17.46, from \$1.49 (Qwen3-32B) to \$42.20 (Opus~4.6). This equals the estimated model token cost of serving about 322 SOP-Bench tasks under the observed task mix (Appendix~\ref{app:training_cost}). Table~\ref{tab:train_vs_infer} gives the per-model values next to the inference cost of serving one million SOP-Bench tasks with the adapted agent. Macro updates stop early: 91.5\% of runs accepted none after iteration 3, and 96.4\% accepted none after iteration 4. Figure~\ref{fig:convergence} shows the best validation accuracy after each iteration.

\begin{table}[!htb]
\centering
\footnotesize
\setlength{\tabcolsep}{4pt}
\begin{tabular}{@{}lrrrr@{}}
\toprule
\textbf{Model} & \textbf{Run (\$)} & \textbf{1M tasks (\$)} & \textbf{Tasks} & \textbf{Share} \\
\midrule
Qwen3-32B     & 1.49  & 3{,}142   & 476 & 0.048\% \\
DeepSeek v3.2 & 6.98  & 19{,}157  & 364 & 0.036\% \\
Haiku 4.5     & 6.76  & 21{,}052  & 321 & 0.032\% \\
Sonnet 4.6    & 29.85 & 94{,}176  & 317 & 0.032\% \\
Opus 4.6      & 42.20 & 133{,}655 & 316 & 0.032\% \\
\midrule
\textit{Mean} & 17.46 & 54{,}236  & 322 & 0.032\% \\
\bottomrule
\end{tabular}
\caption{One-time adaptation cost of one ARISE-Teacher run and inference cost of serving one million SOP-Bench tasks with the adapted agent. \emph{Tasks}: served tasks whose inference cost equals one run. \emph{Share}: run cost as a percentage of serving 1M tasks. Values are means over 33 runs per model (11 domains, three runs). The mean row uses the 5-model mean costs.}
\label{tab:train_vs_infer}
\end{table}

\subsection{Case Studies}
\label{sec:cases}
\begin{figure*}[!t]
\begin{tcbraster}[raster columns=2, raster equal height, raster column skip=6pt]
\begin{tcolorbox}[casebox, title={\textbf{Case 1: micro memory} (dangerous goods, test task 242)}]
\caselabel{Input} sds\_label=5, handling=4, transportation=0, disposal=0. \textrm{Expected:} Unable to Decide\\[3pt]
\caselabel{Before} \textrm{(retrieval off)}: this is exactly two, not more than two, so we proceed with imputation. [\ldots] \hlbad{Hazard Class D}\\[3pt]
\caselabel{Retrieved correction} \textrm{(from training task 111)}: [WHEN: exactly two out of four component scores are missing/zero (e.g., handling\_score=0 and disposal\_score=0)] classify as "unable to decide" [\ldots]\\[3pt]
\caselabel{After}: Since exactly two scores are missing/zero, this meets the threshold for "Unable to Decide" [\ldots] \hlgood{Unable to Decide}
\end{tcolorbox}
\begin{tcolorbox}[casebox, title={\textbf{Case 2: macro memory} (customer service, test task 126)}]
\caselabel{Input} account TERMINATED, reason 'Contract expired'. \textrm{Expected:} FAILED\\[3pt]
\caselabel{Before} \textrm{(baseline)}: terminated accounts are ineligible for support services. [\ldots] Final status: \hlbad{RESOLVED}\\[3pt]
\caselabel{Macro rule}: [WHEN: account status is SUSPENDED or TERMINATED and the issue cannot be resolved through normal support actions (e.g., [\ldots] contract expired)] Output "failed" as the case outcome [\ldots]\\[3pt]
\caselabel{After} \textrm{(macro rules only)}: terminated accounts are ineligible for any support actions. [\ldots] Final case outcome: \hlgood{FAILED}
\end{tcolorbox}
\end{tcbraster}
\caption{Case studies with Opus~4.6. ARISE outputs come from run~3. \emph{Before} is the same run with retrieval switched off in Case~1 and the baseline in Case~2. Red marks the wrong answer and green the correct one. Excerpts are verbatim, and [\ldots] marks cuts.}
\label{fig:cases}
\end{figure*}
Figure~\ref{fig:cases} shows one test case per memory level, before and after adaptation.
\paragraph{Case 1: micro memory.} The SOP imputes missing scores unless more than two are missing. The labels give \texttt{Unable to Decide} already for two. No macro rule covers this, and only training task~111 has two missing scores. Its correction is retrieved for all four test tasks with two missing scores, though each misses a different pair. These go from 0 to 4 correct, and all five baselines miss them. The nine tasks with one missing score also retrieve it, and none is answered \texttt{Unable to Decide}. Three other runs learn a similar correction and get three or four right.
\paragraph{Case 2: macro memory.} The SOP concludes a case for an ineligible account but names no status. The labels use \texttt{FAILED}. On nine such test tasks, the five baselines answer 23 of 45 correctly, mostly erring with \texttt{PENDING\_ACTION}. The consolidated rule names the account states. With macro rules only, this run answers all nine, and the 15 ARISE-Teacher runs answer 130 of 135. Task~152 has a lifted suspension, and this run still escalates it correctly. Three of the 15 runs close it as \texttt{FAILED}.
\subsection{Hyperparameter Sensitivity}
\label{sec:sensitivity}

We vary one setting at a time around the default and report the SOP-Bench 5-model average (Table~\ref{tab:sensitivity}). Each variant is a single training run, so small differences from the three-run default should be interpreted cautiously. The default (87.4) is the mean of three runs whose 5-model averages are 87.2, 87.8 and 87.2 ($\sigma$ 0.3\,pp). Batch size, split, budget and $\epsilon_0$ stay within 1.1\,pp of the default. Top-3 retrieval at $\theta{=}0.8$ scores 85.6 and top-10 scores 88.2.

\begin{table}[!htb]
\centering
\footnotesize
\setlength{\tabcolsep}{3pt}
\begin{tabular}{@{}lccc@{}}
\toprule
\textbf{Hyperparameter} & \textbf{Low} & \textbf{Def.} & \textbf{High} \\
\midrule
Retrieval policy $(k,\theta)$               & 85.6 & \textbf{87.4} & 88.2 \\
Batch size $B$ (10\,/\,30\,/\,50)                  & 86.3 & \textbf{87.4} & 87.6 \\
Train/val split (60/40\,/\,70/30\,/\,80/20)        & 87.0 & \textbf{87.4} & 88.1 \\
Budget $k_{\max}{\times}B$ (3$\times$50\,/\,5$\times$30\,/\,15$\times$10) & 87.3 & \textbf{87.4} & 87.6 \\
Macro threshold $\epsilon_0$ (1\,/\,2\,/\,5)       & 88.5 & \textbf{87.4} & 87.7 \\
\bottomrule
\end{tabular}
\caption{SOP-Bench accuracy (\%) under one-at-a-time changes (low\,/\,default\,/\,high). Retrieval uses $\theta{=}0.8$ at $k{=}3$ and $\theta{=}0.6$ otherwise. Bold marks the default configuration rather than the best result.}
\label{tab:sensitivity}
\end{table}

\section{Guidance for Enterprise Deployment}
\label{sec:guidance}

\paragraph{What drives the gains.} Consolidated rules provide most of the gain. Retrieved corrections add 0.7\,pp and supply source material for rule learning. ARISE-Self retains a 10.0\,pp gain, so the same model can attribute and consolidate. The teacher adds 0.9\,pp on SOP-Bench and 3.2\,pp on $\tau$, which suggests more value in workflows that span conversational turns and tools.

\paragraph{Choosing the attributor.} Self-attribution needs no extra model and keeps most of the SOP-Bench gain. A teacher helps most on multi-turn tasks.

\paragraph{Validation behavior.} The gate keeps the current rules by default and accepted no update in 32\% of SOP-Bench training runs. Accepted rule sets hold 3.4 rules on average (Figure~\ref{fig:cases} shows one). The gate checks validation accuracy only; safety and policy checks remain separate.

\paragraph{Choosing the deployment model.} With ARISE-Teacher, Sonnet~4.6 achieves higher average accuracy than the Opus~4.6 baseline on both benchmarks, at 33\% and 39\% lower inference cost per task, respectively (Table~\ref{tab:efficiency_all}). Haiku~4.5 also exceeds the Sonnet~4.6 baseline on SOP-Bench. A cheaper model with memory can thus beat a stronger model without it.

\paragraph{Maintenance.} When a procedure or deployment model changes, an ARISE rerun rebuilds the memories, and the tools and base prompt stay unchanged. For Haiku~4.5 on SOP-Bench, a rerun costs an estimated \$6.76, against \$21,052 for serving 1M tasks.

\section{Conclusion}
\label{sec:conclusion}
ARISE is a practical memory architecture for accurate and cost-efficient enterprise agents. It turns failed tasks into scoped corrections and a compact set of macro rules, accepting rule updates only when they pass a held-out validation gate. ARISE-Teacher improves over the baseline by 10.9\,pp on SOP-Bench and 5.8\,pp on $\tau$ and reduces model calls by 4.0\% on average. With ARISE-Teacher, Sonnet~4.6 beats the Opus~4.6 baseline on both benchmarks at 33\% and 39\% lower cost per task. Without a stronger model, ARISE-Self improves every model, including small and open-weight models, with an overall gain of 8.4\,pp, within 1.4\,pp of ARISE-Teacher. We hope ARISE helps enterprise agents adapt as their models and procedures evolve. 

\section*{Limitations}

To make ARISE easy to adapt to enterprise agents, we design it to update the injected context through two memory levels, with a compact set of macro rules that an operator can review before release. This design keeps the deployed model and its tools unchanged, so the same loop applies to closed-source models and to new model releases. Future work can explore richer memory structures, such as graph-based memory or more memory levels, and extend the loop to other parts of the agent harness, such as its tools, tool descriptions and workflow logic. ARISE requires verifiable pass or fail outcomes, which are common in enterprise workflows, and extending it to ambiguous tasks without clear outcomes is an important direction for future work. Average gains vary across domains, and some model and domain combinations regress relative to the baseline. Production use requires data governance and memory safety controls that we do not evaluate.

\bibliography{custom}

\clearpage
\appendix
\raggedbottom
\renewcommand{\dbltopfraction}{0.95}
\renewcommand{\dblfloatpagefraction}{0.5}
\setcounter{dbltopnumber}{3}
\renewcommand{\topfraction}{0.95}
\renewcommand{\floatpagefraction}{0.5}
\renewcommand{\textfraction}{0.05}
\setcounter{topnumber}{4}
\setcounter{totalnumber}{6}
\makeatletter\setlength{\@fptop}{0pt}\setlength{\@dblfptop}{0pt}\setlength{\@fpsep}{14pt}\setlength{\@dblfpsep}{14pt}\setlength{\@fpbot}{0pt plus 1fil}\setlength{\@dblfpbot}{0pt plus 1fil}\makeatother

\section{Implementation Details}
\label{app:implementation}

\subsection{Consolidation Prompts}
\label{app:prompts}

Consolidation receives the corrections from the current batch. Figure~\ref{fig:prompts} gives the attributor prompt.

\begin{tcolorbox}[colback=gray!3, colframe=gray!50!black, title={\textbf{Consolidation prompt: append candidate}}, fonttitle=\small, breakable]
\small\ttfamily\raggedright
You are adding rules to an AI agent's system prompt to fix specific failures.\\[4pt]
Current rules:\\
\{current\_rules or ``(none yet)''\}\\[4pt]
New failure patterns observed:\\
\{accuracy\_lessons\}\\[4pt]
Generate 1-2 NEW rules. CRITICAL REQUIREMENTS:\\
1. Each rule MUST start with a condition specifying WHEN it applies: ``[WHEN: <specific task type or situation>] <rule>''. This prevents the rule from interfering with unrelated tasks.\\
2. Each rule must fix a SPECIFIC correctness issue (wrong tool, wrong argument)\\
3. NEVER write generic rules that apply to ALL tasks\\
4. NEVER write generic rules about pacing or caution\\
5. Each rule must ONLY activate for the relevant condition.\\[4pt]
Output ONLY the 1-2 new rules (no explanation, no numbering):
\end{tcolorbox}

\begin{tcolorbox}[colback=gray!3, colframe=gray!50!black, title={\textbf{Consolidation prompt: rewrite candidate}}, fonttitle=\small, breakable]
\small\ttfamily\raggedright
You are writing the COMPLETE set of conditional rules for an AI agent.\\[4pt]
Failure patterns observed:\\
\{accuracy\_lessons\}\\[4pt]
Write 3-5 CONDITIONAL rules. STRICT FORMAT:\\
Each rule MUST follow this pattern:\\
~~[WHEN: <specific task type or situation>] <rule>\\[4pt]
REQUIREMENTS:\\
\textbullet{} Every rule must have a [WHEN: ...] condition, NO universal rules\\
\textbullet{} Each rule targets a SPECIFIC failure pattern from the observations above\\
\textbullet{} NEVER write generic rules about pacing or caution\\
\textbullet{} Rules must not interfere with tasks where they do not apply\\[4pt]
Output ONLY the rules (one per line):
\end{tcolorbox}

\subsection{Configuration}
\label{app:config}

\begin{table}[h]
\centering
\footnotesize
\hyphenpenalty=10000
\exhyphenpenalty=10000
\setlength{\tabcolsep}{3pt}
\caption{Settings used in the reported runs.}
\label{tab:verified_configuration}
\begin{tabular}{@{}>{\raggedright\arraybackslash}p{0.23\columnwidth}>{\raggedright\arraybackslash}p{0.73\columnwidth}@{}}
\toprule
\textbf{Setting} & \textbf{Value} \\
\midrule
Models & Opus~4.6, Sonnet~4.6, Haiku~4.5, DeepSeek~v3.2 and Qwen3-32B via Amazon Bedrock \\
Temperature & 0 for the agent and ARISE, 1.0 for GEPA \\
Simulator & Opus~4.6, temperature 1.0 \\
Data splits & SOP-Bench train and test split by row at 70/30. $\tau^2$ official splits 30/20, 74/40 and 74/40. Training sets shuffled with seed 42, then split 70/30 for iteration and validation \\
Validation size & SOP-Bench 14 to 58, airline 9, retail 23, telecom 23 \\
Memory & 5 macro rules, 5 retrieved corrections, cosine threshold $\theta{=}0.6$ \\
Adaptation & Batch size 30, at most 5 iterations, patience 3, macro margin $\max(2\,\mathrm{pp},50/|\mathcal{D}_{\mathrm{val}}|)$. The micro gate uses the first 20 validation tasks when available with a 5\,pp tolerance. During adaptation, validation scores average three evaluations for telecom and use one elsewhere \\
Step limits & SOP-Bench 10 turns, airline and retail 30 steps, telecom 200 steps \\
Embedding & Cohere Embed English v3, 1,024 dimensions, cosine similarity \\
Pricing & Bedrock US East on demand prices \\
GEPA budget & Opus~4.6, 200 metric calls for SOP-Bench and 100 for $\tau$ \\
\bottomrule
\end{tabular}
\end{table}

\paragraph{Domain coverage.} We use the 11 domains listed in Section~\ref{sec:setup}. Email intent and video annotation are omitted because their released data and scoring interfaces are incompatible with our evaluation.

\section{Additional Results}
\label{app:results}

\subsection{Significance Tests}
\label{app:stats}

Each comparison with Table~\ref{tab:main_results} uses a stratified paired bootstrap over test tasks. Tasks are paired by ID within each domain$\times$model cell and resampled with the same draw for both methods. For $\tau$ evaluations with four trials, each task keeps its trials together. For methods with multiple training runs, each replicate first resamples the reported run units within the cell, then the tasks. The statistic is the value the table prints: a per-model value, a 5-model average or the 11:3 overall average. Accuracy comparisons use 10,000 resamples and efficiency comparisons 5,000. Two sided $p$ values double the smaller empirical tail probability of the uncentered bootstrap difference, use the $+1$ correction, count ties in both tails, and cap at one. We use seed 0 and report unadjusted $p$ values. Table~\ref{tab:stats} gives the accuracy comparisons of the averages.

\begin{table}[h]
\centering
\footnotesize
\setlength{\tabcolsep}{2.5pt}
\caption{Accuracy differences of the averages in Table~\ref{tab:main_results} (A\,/\,B: A minus B in pp), with $p$ values in parentheses.}
\label{tab:stats}
\begin{tabular}{@{}lrrr@{}}
\toprule
\textbf{Comparison} & \textbf{SOP} & \textbf{$\tau$} & \textbf{Overall} \\
\midrule
Teacher\,/\,Base & +10.9 (${<}$.01) & +5.8 (${<}$.01) & +9.8 (${<}$.01) \\
Self\,/\,Base & +10.0 (${<}$.01) & +2.6 (.05) & +8.4 (${<}$.01) \\
Teacher\,/\,GEPA & +1.4 (.02) & +4.5 (${<}$.01) & +2.0 (${<}$.01) \\
\bottomrule
\end{tabular}
\end{table}

\paragraph{Task-weighted averages.} Pooling task outcomes across all cells, thereby weighting domains by test-set size, ARISE-Teacher reaches 88.7 on SOP-Bench against 77.2 for the baseline (+11.5\,pp) and 63.0 on $\tau$ against 56.8 (+6.2\,pp). Against GEPA the pooled differences are +1.9\,pp on SOP-Bench and +4.7\,pp on $\tau$. ARISE-Self reaches 87.9 on SOP-Bench (+10.7\,pp over the baseline).

\paragraph{Micro memory.} To isolate micro memory, every trained run was re-evaluated on its test tasks with its accepted macro rules and micro retrieval switched off. Runs whose test phase retrieved no correction retained their stored result. With the rules fixed, micro memory adds 0.7\,pp on SOP-Bench for ARISE-Teacher (87.4 vs.\ 86.7, fixing 186 test cases and breaking 118) and 0.6\,pp for ARISE-Self.

\subsection{$\tau$ Consistency}
\label{app:tau_consistency}

For a task with $s$ successes among four trials, we compute
\[
\mathrm{pass}^k=\frac{\binom{s}{k}}{\binom{4}{k}},
\qquad
\mathrm{pass}@k=1-\frac{\binom{4-s}{k}}{\binom{4}{k}}.
\]
We average these values over tasks within each run. ARISE-Teacher values then average the three training runs. Table~\ref{tab:tau_consistency_full} reports pass\textasciicircum k for $k{=}1,\dots,4$ and pass@$k$ for $k{=}2,3,4$, averaged over 5 models. Table~\ref{tab:tau_consistency_permodel} gives per-model values for Baseline and ARISE-Teacher.

\begin{table*}[!tp]
\centering
\footnotesize
\setlength{\tabcolsep}{5pt}
\renewcommand{\arraystretch}{0.92}
\caption{Full $\tau$ consistency (\%), averaged over 5 models. pass\textasciicircum k: solved in all $k$ trials; pass@$k$: solved in at least one of $k$ trials. Best per domain in \textbf{bold}.}
\label{tab:tau_consistency_full}
\begin{tabular}{@{}ll|rrrr|rrr@{}}
\toprule
& & \multicolumn{4}{c|}{\textbf{pass\textasciicircum k (all $k$)}} & \multicolumn{3}{c}{\textbf{pass@$k$ ($\geq$1 of $k$)}} \\
\textbf{Domain} & \textbf{Method} & \textbf{$k$=1} & \textbf{2} & \textbf{3} & \textbf{4} & \textbf{2} & \textbf{3} & \textbf{4} \\
\midrule
\multirow{4}{*}{Airline}
 & Baseline & 55.2 & 46.8 & 43.0 & 41.0 & 63.7 & 68.2 & \textbf{71.0} \\
 & GEPA     & 55.2 & 46.8 & 41.8 & 38.0 & 63.7 & 67.0 & 69.0 \\
 & Macro    & 56.8 & 48.0 & 42.2 & 38.0 & 65.5 & 68.5 & 70.0 \\
 & ARISE-T  & \textbf{58.8} & \textbf{51.4} & \textbf{47.2} & \textbf{44.3} & \textbf{66.1} & \textbf{69.2} & \textbf{71.0} \\
\midrule
\multirow{4}{*}{Retail}
 & Baseline & 48.1 & 40.8 & 36.8 & 34.0 & 55.5 & 58.9 & 61.0 \\
 & GEPA     & 46.1 & 36.8 & 31.4 & 28.0 & 55.5 & 59.5 & 61.5 \\
 & Macro    & 47.6 & 39.0 & 34.0 & 30.0 & \textbf{56.2} & \textbf{59.9} & \textbf{62.5} \\
 & ARISE-T  & \textbf{48.5} & \textbf{43.1} & \textbf{40.2} & \textbf{38.0} & 53.9 & 56.4 & 58.0 \\
\midrule
\multirow{4}{*}{Telecom}
 & Baseline & 66.1 & 55.3 & 49.0 & 44.5 & 76.9 & 81.4 & 84.0 \\
 & GEPA     & 71.8 & 62.3 & 56.4 & 52.0 & 81.2 & 84.6 & 86.5 \\
 & Macro    & 74.8 & 62.9 & 55.0 & 49.0 & 86.6 & 90.5 & 92.5 \\
 & ARISE-T  & \textbf{79.5} & \textbf{70.2} & \textbf{64.5} & \textbf{60.7} & \textbf{88.8} & \textbf{92.4} & \textbf{94.2} \\
\bottomrule
\end{tabular}
\end{table*}

\begin{table*}[!tp]
\centering
\footnotesize
\setlength{\tabcolsep}{5pt}
\renewcommand{\arraystretch}{0.92}
\caption{Per-model $\tau$ consistency (\%): Baseline vs.\ ARISE-Teacher on pass\textasciicircum1, pass\textasciicircum4 and pass@4. In the column headings, $p$ abbreviates pass; the last column is the change in pass\textasciicircum4.}
\label{tab:tau_consistency_permodel}
\begin{tabular}{@{}ll|rrr|rrr|r@{}}
\toprule
& & \multicolumn{3}{c|}{\textbf{Baseline}} & \multicolumn{3}{c|}{\textbf{ARISE-Teacher}} & \\
\textbf{Model} & \textbf{Domain} & \textbf{p\textasciicircum1} & \textbf{p\textasciicircum4} & \textbf{p@4} & \textbf{p\textasciicircum1} & \textbf{p\textasciicircum4} & \textbf{p@4} & \textbf{$\Delta$p\textasciicircum4} \\
\midrule
\multirow{3}{*}{Opus 4.6}
 & Airline & 78.8 & 60.0 & 90.0 & 77.5 & 61.7 & 88.3 & $+$1.7 \\
 & Retail  & 67.5 & 62.5 & 75.0 & 66.9 & 64.2 & 69.2 & $+$1.7 \\
 & Telecom & 83.1 & 67.5 & 100.0 & 98.3 & 94.2 & 100.0 & $+$26.7 \\
\midrule
\multirow{3}{*}{Sonnet 4.6}
 & Airline & 81.2 & 75.0 & 90.0 & 77.5 & 68.3 & 86.7 & $-$6.7 \\
 & Retail  & 58.8 & 45.0 & 70.0 & 60.2 & 51.7 & 69.2 & $+$6.7 \\
 & Telecom & 83.8 & 67.5 & 95.0 & 94.2 & 81.7 & 100.0 & $+$14.2 \\
\midrule
\multirow{3}{*}{Haiku 4.5}
 & Airline & 55.0 & 30.0 & 85.0 & 64.6 & 43.3 & 76.7 & $+$13.3 \\
 & Retail  & 55.0 & 30.0 & 77.5 & 57.9 & 43.3 & 70.0 & $+$13.3 \\
 & Telecom & 50.6 & 32.5 & 62.5 & 71.2 & 45.0 & 95.0 & $+$12.5 \\
\midrule
\multirow{3}{*}{DeepSeek v3.2}
 & Airline & 41.2 & 30.0 & 55.0 & 50.4 & 36.7 & 60.0 & $+$6.7 \\
 & Retail  & 39.4 & 27.5 & 50.0 & 38.1 & 25.8 & 52.5 & $-$1.7 \\
 & Telecom & 83.1 & 52.5 & 100.0 & 89.0 & 69.2 & 99.2 & $+$16.7 \\
\midrule
\multirow{3}{*}{Qwen3-32B}
 & Airline & 20.0 & 10.0 & 35.0 & 23.8 & 11.7 & 43.3 & $+$1.7 \\
 & Retail  & 20.0 & 5.0 & 32.5 & 19.6 & 5.0 & 29.2 & $+$0.0 \\
 & Telecom & 30.0 & 2.5 & 62.5 & 44.6 & 13.3 & 76.7 & $+$10.8 \\
\bottomrule
\end{tabular}
\end{table*}

\subsection{Component Variants}
\label{app:ablation}

Figure~\ref{fig:ablation} reports three variants. \emph{w/o micro memory} reevaluates every ARISE-Teacher run with its accepted rules and micro retrieval switched off. \emph{w/o attribution} stores each failed trace with its expected output instead of a correction. \emph{w/o teacher} is ARISE-Self, in which the deployment model attributes and consolidates.

\subsection{Adaptation versus Inference Cost}
\label{app:training_cost}

Adaptation cost includes initial baseline validation, training batches, candidate validation, final micro gate validation, attribution and consolidation. It excludes embedding calls and final testing. Agent executions use each run's test phase cost per task, while attribution and consolidation use observed input tokens and the maximum output token budget. Across models, one run costs \$1.49 to \$42.20, with a mean of \$17.46, equivalent to 316 to 476 served tasks and 322 at the 5 model mean.

\subsection{Inference Efficiency}
\label{app:efficiency}

Table~\ref{tab:efficiency_all} reports per-task input tokens, output tokens, model calls and cost for every model and method. Per-task cost is $(\text{input tokens}/10^6)P_{\mathrm{in}}+(\text{output tokens}/10^6)P_{\mathrm{out}}$ with the prices in \S\ref{sec:setup}. On SOP-Bench, the 5-model mean cost per task is \$0.0548 for the baseline, \$0.0539 for ARISE-Self and \$0.0542 for ARISE-Teacher. Table~\ref{tab:main_results} reports the mean of the 5 per-model changes, computed from unrounded values.

\begin{table*}[!t]
\centering
\small
\setlength{\tabcolsep}{4pt}
\caption{Per-task efficiency: average input tokens, output tokens, model calls and cost (\$) per model and method. SOP-Bench is averaged over 11 domains and $\tau$ over 3 domains. GEPA's SOP-Bench values come from the test pass of its own run. On Opus~4.6 the ARISE-Self row repeats ARISE-Teacher.}
\label{tab:efficiency_all}
\begin{tabular}{@{}ll|rrrr|rrrr@{}}
\toprule
& & \multicolumn{4}{c|}{\textbf{SOP-Bench}} & \multicolumn{4}{c}{\textbf{$\tau$}} \\
\textbf{Method} & \textbf{Model} & \textbf{InTok} & \textbf{OutTok} & \textbf{Calls} & \textbf{Cost} & \textbf{InTok} & \textbf{OutTok} & \textbf{Calls} & \textbf{Cost} \\
\midrule
 \multirow{5}{*}{Baseline} & Qwen3-32B     & 21{,}105 & 333 & 5.28 & 0.0034 & 135{,}231 & 1{,}260 & 16.74 & 0.0210 \\
  & DeepSeek v3.2 & 25{,}695 & 1{,}121 & 6.06 & 0.0180 & 119{,}780 & 2{,}267 & 14.25 & 0.0785 \\
  & Haiku 4.5     & 13{,}570 & 1{,}141 & 2.89 & 0.0212 & 113{,}758 & 2{,}106 & 12.06 & 0.1367 \\
  & Sonnet 4.6    & 20{,}461 & 1{,}479 & 3.96 & 0.0919 & 103{,}453 & 1{,}529 & 11.69 & 0.3666 \\
  & Opus 4.6      & 19{,}595 & 1{,}157 & 3.79 & 0.1396 & 103{,}803 & 1{,}651 & 11.36 & 0.6163 \\
\midrule
 \multirow{5}{*}{ExpeL} & Qwen3-32B     & 22{,}236 & 332 & 5.42 & 0.0035 & 141{,}931 & 1{,}126 & 17.28 & 0.0220 \\
  & DeepSeek v3.2 & 26{,}545 & 1{,}134 & 5.95 & 0.0186 & 131{,}635 & 2{,}216 & 15.09 & 0.0857 \\
  & Haiku 4.5     & 14{,}454 & 1{,}190 & 2.95 & 0.0224 & 94{,}485 & 1{,}895 & 10.47 & 0.1144 \\
  & Sonnet 4.6    & 21{,}086 & 1{,}554 & 3.89 & 0.0952 & 94{,}836 & 1{,}454 & 10.56 & 0.3369 \\
  & Opus 4.6      & 20{,}168 & 1{,}216 & 3.69 & 0.1444 & 104{,}654 & 1{,}640 & 11.03 & 0.6207 \\
\midrule
 \multirow{5}{*}{Reflexion} & Qwen3-32B     & 22{,}292 & 334 & 5.40 & 0.0035 & 152{,}999 & 1{,}361 & 17.86 & 0.0238 \\
  & DeepSeek v3.2 & 26{,}822 & 1{,}126 & 6.08 & 0.0187 & 122{,}657 & 2{,}266 & 14.22 & 0.0802 \\
  & Haiku 4.5     & 15{,}056 & 1{,}163 & 3.07 & 0.0230 & 105{,}005 & 2{,}090 & 11.21 & 0.1270 \\
  & Sonnet 4.6    & 20{,}481 & 1{,}410 & 3.90 & 0.0908 & 98{,}030 & 1{,}373 & 11.06 & 0.3461 \\
  & Opus 4.6      & 19{,}373 & 1{,}130 & 3.70 & 0.1376 & 116{,}948 & 1{,}664 & 12.47 & 0.6890 \\
\midrule
 \multirow{5}{*}{GEPA} & Qwen3-32B     & 16{,}618 & 248 & 4.20 & 0.0026 & 126{,}442 & 1{,}128 & 15.45 & 0.0196 \\
  & DeepSeek v3.2 & 26{,}327 & 1{,}088 & 5.88 & 0.0183 & 126{,}112 & 2{,}202 & 14.20 & 0.0823 \\
  & Haiku 4.5     & 14{,}203 & 1{,}086 & 2.82 & 0.0216 & 95{,}109 & 1{,}883 & 10.55 & 0.1150 \\
  & Sonnet 4.6    & 21{,}039 & 1{,}323 & 3.89 & 0.0913 & 112{,}433 & 1{,}511 & 11.93 & 0.3960 \\
  & Opus 4.6      & 19{,}146 & 1{,}121 & 3.61 & 0.1361 & 114{,}313 & 1{,}691 & 11.73 & 0.6752 \\
\midrule
 \multirow{5}{*}{Macro} & Qwen3-32B     & 18{,}396 & 283 & 4.75 & 0.0029 & 150{,}155 & 1{,}446 & 17.86 & 0.0234 \\
  & DeepSeek v3.2 & 26{,}314 & 1{,}175 & 6.03 & 0.0185 & 113{,}099 & 2{,}178 & 13.57 & 0.0742 \\
  & Haiku 4.5     & 13{,}690 & 1{,}060 & 2.86 & 0.0209 & 110{,}927 & 2{,}266 & 11.70 & 0.1345 \\
  & Sonnet 4.6    & 20{,}545 & 1{,}525 & 3.84 & 0.0930 & 111{,}477 & 1{,}499 & 12.17 & 0.3926 \\
  & Opus 4.6      & 19{,}027 & 1{,}147 & 3.63 & 0.1362 & 106{,}030 & 1{,}803 & 11.25 & 0.6327 \\
\midrule
 \multirow{5}{*}{ARISE-Self} & Qwen3-32B     & 19{,}561 & 297 & 4.82 & 0.0031 & 123{,}172 & 1{,}083 & 15.65 & 0.0191 \\
  & DeepSeek v3.2 & 25{,}981 & 1{,}128 & 5.94 & 0.0182 & 113{,}840 & 2{,}046 & 13.81 & 0.0744 \\
  & Haiku 4.5     & 14{,}451 & 1{,}105 & 2.91 & 0.0220 & 108{,}659 & 2{,}034 & 11.73 & 0.1307 \\
  & Sonnet 4.6    & 20{,}926 & 1{,}440 & 3.87 & 0.0928 & 103{,}618 & 1{,}437 & 11.65 & 0.3656 \\
  & Opus 4.6      & 18{,}729 & 1{,}114 & 3.51 & 0.1337 & 105{,}655 & 1{,}672 & 11.40 & 0.6271 \\
\midrule
 \multirow{5}{*}{ARISE-T} & Qwen3-32B     & 19{,}827 & 280 & 4.83 & 0.0031 & 119{,}508 & 1{,}014 & 15.02 & 0.0185 \\
  & DeepSeek v3.2 & 27{,}471 & 1{,}148 & 6.00 & 0.0192 & 117{,}207 & 2{,}142 & 13.89 & 0.0766 \\
  & Haiku 4.5     & 13{,}755 & 1{,}077 & 2.82 & 0.0211 & 110{,}278 & 2{,}037 & 11.62 & 0.1325 \\
  & Sonnet 4.6    & 21{,}266 & 1{,}454 & 3.88 & 0.0942 & 106{,}896 & 1{,}455 & 11.76 & 0.3768 \\
  & Opus 4.6      & 18{,}729 & 1{,}114 & 3.51 & 0.1337 & 105{,}655 & 1{,}672 & 11.40 & 0.6271 \\
\bottomrule
\end{tabular}
\end{table*}

\subsection{Full Results}
\label{app:full_results}

Table~\ref{tab:raw_sop} reports SOP-Bench accuracy for every model, domain and method; Table~\ref{tab:raw_tau} does the same for $\tau$.

\begin{table*}[p]
\centering
\small
\setlength{\tabcolsep}{3pt}
\caption{Full SOP-Bench accuracy (\%) per model, domain and method. ARISE-Teacher (ARISE-T) and ARISE-Self (Self) average three training runs, and on Opus~4.6 the Self cells repeat ARISE-T. Baseline, ExpeL, Reflexion, GEPA and Macro use one run. GEPA is the per-task re-evaluation of its best prompt (\S\ref{sec:setup}).}
\label{tab:raw_sop}
\begin{tabular}{@{}ll|rrrrrrr@{}}
\toprule
\textbf{Domain} & \textbf{Model} & \textbf{Baseline} & \textbf{ExpeL} & \textbf{Reflexion} & \textbf{GEPA} & \textbf{Macro} & \textbf{Self} & \textbf{ARISE-T} \\
\midrule
\multirow{5}{*}{Aircraft Insp.} & Opus 4.6 & 97.1 & 88.2 & 97.1 & 97.1 & 97.1 & 96.9 & 96.9 \\
 & Sonnet 4.6 & 67.6 & 97.1 & 67.6 & 94.1 & 97.1 & 97.1 & 97.0 \\
 & Haiku 4.5 & 97.1 & 97.1 & 97.1 & 97.1 & 97.1 & 97.1 & 97.1 \\
 & DeepSeek v3.2 & 79.4 & 73.5 & 85.3 & 94.1 & 91.2 & 85.3 & 86.3 \\
 & Qwen3-32B & 97.1 & 97.1 & 97.1 & 94.1 & 97.1 & 96.1 & 95.1 \\
\midrule
\multirow{5}{*}{Content Flag.} & Opus 4.6 & 37.3 & 62.7 & 43.2 & 92.2 & 94.1 & 92.8 & 92.8 \\
 & Sonnet 4.6 & 37.3 & 41.2 & 45.1 & 70.6 & 82.4 & 87.6 & 86.3 \\
 & Haiku 4.5 & 7.8 & 11.8 & 7.8 & 80.4 & 78.4 & 81.7 & 77.8 \\
 & DeepSeek v3.2 & 51.1 & 49.0 & 52.9 & 86.3 & 72.5 & 64.7 & 88.2 \\
 & Qwen3-32B & 0.0 & 0.0 & 0.0 & 11.8 & 43.1 & 49.7 & 49.7 \\
\midrule
\multirow{5}{*}{Customer Svc.} & Opus 4.6 & 87.2 & 100.0 & 89.4 & 97.9 & 97.9 & 98.6 & 98.6 \\
 & Sonnet 4.6 & 91.5 & 93.6 & 93.6 & 93.6 & 100.0 & 93.6 & 100.0 \\
 & Haiku 4.5 & 87.2 & 87.2 & 87.2 & 100.0 & 100.0 & 100.0 & 100.0 \\
 & DeepSeek v3.2 & 46.8 & 85.1 & 38.3 & 97.9 & 100.0 & 81.6 & 97.9 \\
 & Qwen3-32B & 40.4 & 51.1 & 31.9 & 66.0 & 95.7 & 85.8 & 71.6 \\
\midrule
\multirow{5}{*}{Dangerous Gds.} & Opus 4.6 & 74.7 & 85.5 & 75.9 & 90.4 & 90.4 & 95.6 & 95.6 \\
 & Sonnet 4.6 & 74.7 & 74.7 & 74.7 & 95.2 & 81.9 & 93.6 & 89.6 \\
 & Haiku 4.5 & 86.7 & 88.0 & 85.5 & 89.2 & 95.2 & 92.0 & 92.0 \\
 & DeepSeek v3.2 & 69.9 & 68.7 & 68.7 & 69.9 & 71.1 & 86.3 & 93.6 \\
 & Qwen3-32B & 56.6 & 60.2 & 67.5 & 74.7 & 50.6 & 88.4 & 90.0 \\
\midrule
\multirow{5}{*}{Know Your Biz.} & Opus 4.6 & 60.7 & 50.0 & 60.7 & 82.1 & 78.6 & 83.3 & 83.3 \\
 & Sonnet 4.6 & 60.7 & 60.7 & 60.7 & 57.1 & 57.1 & 57.1 & 58.3 \\
 & Haiku 4.5 & 57.1 & 53.6 & 50.0 & 53.6 & 53.6 & 53.6 & 54.8 \\
 & DeepSeek v3.2 & 60.7 & 60.7 & 60.7 & 78.6 & 71.4 & 59.5 & 63.1 \\
 & Qwen3-32B & 53.6 & 53.6 & 53.6 & 53.6 & 46.4 & 48.8 & 53.6 \\
\midrule
\multirow{5}{*}{Patient Intake} & Opus 4.6 & 85.0 & 85.0 & 85.0 & 85.0 & 85.0 & 84.9 & 84.9 \\
 & Sonnet 4.6 & 85.0 & 20.0 & 80.0 & 85.0 & 80.0 & 80.0 & 86.2 \\
 & Haiku 4.5 & 85.0 & 80.0 & 85.0 & 85.0 & 85.0 & 85.0 & 85.0 \\
 & DeepSeek v3.2 & 77.8 & 85.0 & 85.0 & 85.0 & 85.0 & 85.0 & 85.0 \\
 & Qwen3-32B & 85.0 & 85.0 & 85.0 & 85.0 & 85.0 & 85.0 & 85.0 \\
\midrule
\multirow{5}{*}{Referral v1} & Opus 4.6 & 96.7 & 100.0 & 96.7 & 98.3 & 100.0 & 98.3 & 98.3 \\
 & Sonnet 4.6 & 91.7 & 98.3 & 96.7 & 95.0 & 96.7 & 100.0 & 98.9 \\
 & Haiku 4.5 & 95.0 & 95.0 & 96.7 & 95.0 & 100.0 & 99.4 & 100.0 \\
 & DeepSeek v3.2 & 98.3 & 95.0 & 96.7 & 96.7 & 95.0 & 97.2 & 98.9 \\
 & Qwen3-32B & 90.0 & 90.0 & 100.0 & 93.3 & 98.3 & 100.0 & 99.4 \\
\midrule
\multirow{5}{*}{Referral v2} & Opus 4.6 & 96.7 & 95.0 & 96.7 & 96.7 & 96.7 & 96.6 & 96.6 \\
 & Sonnet 4.6 & 95.0 & 95.0 & 96.7 & 91.7 & 95.0 & 95.6 & 95.6 \\
 & Haiku 4.5 & 90.0 & 90.0 & 90.0 & 93.3 & 88.3 & 86.1 & 90.0 \\
 & DeepSeek v3.2 & 96.7 & 96.7 & 95.0 & 96.7 & 98.3 & 96.1 & 96.1 \\
 & Qwen3-32B & 35.0 & 43.3 & 36.7 & 48.3 & 40.0 & 46.1 & 40.0 \\
\midrule
\multirow{5}{*}{Traffic Spoof.} & Opus 4.6 & 86.7 & 60.0 & 86.7 & 86.7 & 86.7 & 86.7 & 86.7 \\
 & Sonnet 4.6 & 86.7 & 86.7 & 86.7 & 86.7 & 81.7 & 86.7 & 79.4 \\
 & Haiku 4.5 & 86.7 & 85.0 & 86.7 & 86.7 & 86.7 & 75.0 & 86.7 \\
 & DeepSeek v3.2 & 85.0 & 68.3 & 81.7 & 86.7 & 85.0 & 86.1 & 75.0 \\
 & Qwen3-32B & 86.7 & 86.7 & 86.7 & 86.7 & 86.7 & 84.4 & 86.7 \\
\midrule
\multirow{5}{*}{Video Class.} & Opus 4.6 & 95.0 & 98.3 & 96.7 & 96.7 & 98.3 & 97.2 & 97.2 \\
 & Sonnet 4.6 & 98.3 & 75.0 & 98.3 & 98.3 & 96.7 & 98.3 & 96.1 \\
 & Haiku 4.5 & 98.3 & 91.7 & 98.3 & 98.3 & 98.3 & 95.6 & 95.6 \\
 & DeepSeek v3.2 & 95.0 & 96.7 & 96.7 & 96.7 & 96.7 & 96.1 & 97.2 \\
 & Qwen3-32B & 98.3 & 98.3 & 96.7 & 96.7 & 96.7 & 98.3 & 98.3 \\
\midrule
\multirow{5}{*}{Warehouse Insp.} & Opus 4.6 & 82.2 & 82.2 & 82.2 & 93.3 & 93.3 & 93.3 & 93.3 \\
 & Sonnet 4.6 & 82.2 & 82.2 & 82.2 & 93.3 & 93.3 & 93.3 & 93.3 \\
 & Haiku 4.5 & 80.0 & 42.2 & 82.2 & 93.3 & 91.1 & 93.3 & 88.9 \\
 & DeepSeek v3.2 & 82.2 & 82.2 & 82.2 & 93.3 & 93.3 & 89.6 & 93.3 \\
 & Qwen3-32B & 82.2 & 84.4 & 82.2 & 91.1 & 84.4 & 80.0 & 89.6 \\
\bottomrule
\end{tabular}
\end{table*}

\begin{table*}[!tp]
\centering
\footnotesize
\setlength{\tabcolsep}{4pt}
\renewcommand{\arraystretch}{0.92}
\caption{Full $\tau$ accuracy (\%) per model, domain and method. ARISE-Self uses one pass from each of three runs, except Opus~4.6, which reuses the identical Teacher evaluations. Other methods use four trials, with ARISE-Teacher averaged over three runs.}
\label{tab:raw_tau}
\begin{tabular}{@{}ll|rrrrrrr@{}}
\toprule
\textbf{Domain} & \textbf{Model} & \textbf{Baseline} & \textbf{ExpeL} & \textbf{Reflexion} & \textbf{GEPA} & \textbf{Macro} & \textbf{Self} & \textbf{ARISE-T} \\
\midrule
\multirow{5}{*}{Airline}
 & Opus 4.6 & 78.8 & 78.8 & 75.0 & 75.0 & 73.8 & 77.5 & 77.5 \\
 & Sonnet 4.6 & 81.2 & 78.8 & 75.0 & 77.5 & 72.5 & 76.7 & 77.5 \\
 & Haiku 4.5 & 55.0 & 61.2 & 62.5 & 56.2 & 62.5 & 51.7 & 64.6 \\
 & DeepSeek v3.2 & 41.2 & 48.8 & 52.5 & 50.0 & 48.8 & 46.7 & 50.4 \\
 & Qwen3-32B & 20.0 & 23.8 & 20.0 & 17.5 & 26.2 & 18.3 & 23.8 \\
\midrule
\multirow{5}{*}{Retail}
 & Opus 4.6 & 67.5 & 65.0 & 63.8 & 62.5 & 64.4 & 66.9 & 66.9 \\
 & Sonnet 4.6 & 58.8 & 58.1 & 62.5 & 63.1 & 59.4 & 60.0 & 60.2 \\
 & Haiku 4.5 & 55.0 & 56.2 & 53.1 & 58.1 & 60.0 & 59.2 & 57.9 \\
 & DeepSeek v3.2 & 39.4 & 36.2 & 37.5 & 32.5 & 36.9 & 38.3 & 38.1 \\
 & Qwen3-32B & 20.0 & 24.4 & 14.4 & 14.4 & 17.5 & 20.0 & 19.6 \\
\midrule
\multirow{5}{*}{Telecom}
 & Opus 4.6 & 83.1 & 85.6 & 81.9 & 95.0 & 94.4 & 98.3 & 98.3 \\
 & Sonnet 4.6 & 83.8 & 65.6 & 81.9 & 89.4 & 89.4 & 90.0 & 94.2 \\
 & Haiku 4.5 & 50.6 & 43.8 & 46.9 & 50.6 & 57.5 & 59.2 & 71.2 \\
 & DeepSeek v3.2 & 83.1 & 81.2 & 80.6 & 86.9 & 81.2 & 80.8 & 89.0 \\
 & Qwen3-32B & 30.0 & 25.6 & 30.6 & 36.9 & 51.3 & 43.3 & 44.6 \\
\bottomrule
\end{tabular}
\end{table*}

\end{document}